# *Can LLMs Solve My Grandma's Riddle?* Evaluating Multilingual Large Language Models on Reasoning Traditional Bangla Tricky Riddles

**Nurul Labib Sayeedi[1]\*, Md. Faiyaz Abdullah Sayeedi[1, 3]\*, Khushnur Binte Jahangir[1], Swakkhar Shatabda[2], Sarah M. Preum[4]**

[1]United International University, Bangladesh   [2]BRAC University, Bangladesh
[3] Center for Computational & Data Sciences, Independent University, Bangladesh,
[4] Dartmouth College, USA

`nsayeedi2410045@bsds.uiu.ac.bd`, `msayeedi212049@bscse.uiu.ac.bd`, `khushnur@cse.uiu.ac.bd`
`swakkhar.shatabda@bracu.ac.bd`, `sarah.masud.preum@dartmouth.edu`

## Abstract

Large Language Models (LLMs) show impressive performance on many NLP benchmarks, yet their ability to reason in figurative, culturally grounded, and low-resource settings remains underexplored. We address this gap for Bangla by introducing BANGLARIDDLEEVAL, a benchmark of 1,244 traditional Bangla riddles instantiated across four tasks (**4,976** riddle–task artifacts in total). Using an LLM-based pipeline, we generate Chain-of-Thought explanations, semantically coherent distractors, and fine-grained ambiguity annotations, and evaluate a diverse suite of open-source and closed-source models under different prompting strategies. Models achieve moderate semantic overlap on generative QA but low correctness, MCQ accuracy peaks at only about 56% versus an 83% human baseline, and ambiguity resolution ranges from roughly 26% to 68%, with high-quality explanations confined to the strongest models. These results show that current LLMs capture some cues needed for Bangla riddle reasoning but remain far from human-level performance, establishing BANGLARIDDLEEVAL as a challenging new benchmark for low-resource figurative reasoning. All data, code, and evaluation scripts are available on GitHub: https://github.com/Labib1610/BanglaRiddleEval.

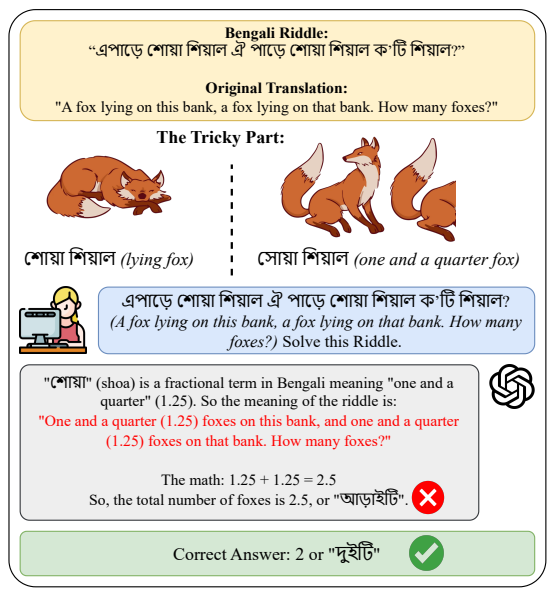

Figure 1: Example of Bangla riddle from BANGLARIDDLEEVAL. The homophone "শোয়া শিয়াল" ("lying fox" vs. "one-and-a-quarter fox") tricks LLM into answering 2.5 foxes, while the correct answer is 2, illustrating the challenge of traditional Bengali riddle reasoning.

## 1 Introduction

Large Language Models have achieved impressive performance across a wide range of natural language understanding and generation tasks, from question answering and summarization to code generation and multi-turn dialogue (Dong et al., 2022; Karanikolas et al., 2023). Despite these advances, it remains unclear how well such models truly reason beyond pattern matching, especially when language is figurative, culturally grounded, and low-resource (Azime et al., 2025). Riddles offer a particularly challenging testbed: they are short, deceptively simple texts that require interpreting metaphors, resolving ambiguity, and connecting commonsense knowledge to implicit clues rather than relying on surface-level cues (Manna et al., 2024).

Existing evaluations of LLMs on reasoning have predominantly focused on high-resource languages (e.g., English, Chinese) (Zhang and Wan, 2022) and structured benchmarks (e.g., math word problems, logical puzzles, and exam-style questions) (Ghosal et al., 2025). By contrast, Bangla, one of the most widely spoken languages in the world, remains underexplored in the context of systematic reasoning evaluation. This gap is especially striking for Bangla riddles, which are deeply embedded in everyday oral tradition and children's literature, and which often encode culturally specific metaphors, idioms, and wordplay that differ substantially from English datasets (Saxena et al., 2025). Riddles thus provide a culturally meaningful probe of Bangla commonsense and figurative reasoning, and a testbed for assessing whether LLMs can handle local narrative forms.

\*Equal Contribution

At the same time, evaluating LLMs on Bangla riddle reasoning presents several practical challenges. First, there is no publicly available standardized Bangla riddles benchmark; existing resources are sparse, often synthetic, or limited to relatively simple tasks (Sefat, 2025). Second, many riddles hinge on subtle "distractor words" and metaphorical triggers, which require fine-grained semantic annotation to turn into well-posed tasks. Third, for a meaningful assessment of reasoning, we need not only final answers but also intermediate explanations of semantic ambiguity. Without such resources, it is difficult to quantify how far current LLMs have progressed towards genuine understanding in Bangla.

In this work, we introduce BANGLARIDDLEEVAL, a new benchmark designed to fill this gap and probe LLM reasoning on Bangla riddles. The benchmark consists of 1,244 unique riddles instantiated across four complementary tasks, yielding a total of 4,976 riddle–task artifacts spanning direct generative QA, reasoning, multiple-choice question answering, and semantic ambiguity resolution. We further benchmark a diverse set of open-source and closed-source LLMs under different prompting regimes and compare them against human baselines. While there has been some prior work on Bangla riddles, BANGLARIDDLEEVAL appears to be among the most comprehensive and human-like evaluations to date, with a particularly challenging and rigorous design for assessing LLM reasoning in Bangla. Our main contributions are:

❶ We construct BANGLARIDDLEEVAL, a dataset of 1,244 unique Bangla riddles. To the best of our knowledge, it is among the first benchmarks to focus specifically on this setting with a systematic, multi-task evaluation.

❷ We frame riddle understanding as four tasks: (i) **Generative QA**, (ii) **Reasoning**, (iii) **Multiple Choice Question**, and (iv) **Semantic Ambiguity Resolution**.

❸ We design an LLM-based pipeline to generate high-fidelity explanations, semantically coherent distractors, and fine-grained semantic ambiguity annotations, with quality controls to reduce trivial shortcuts and annotation artifacts.

❹ We systematically evaluate a suite of open-source and closed-source models under zero-shot, few-shot, and chain-of-thought prompting, and compare them against human performance and an LLM-as-a-Judge assessment.

## 2 Related Work

Early work on riddle-style QA focuses on high-resource languages, primarily English and Chinese. Lin et al. (2021) introduce RiddleSense, a multiple-choice English benchmark targeting figurative, commonsense-rich riddle questions. BiRdQA (Zhang and Wan, 2022) extends this to a bilingual English–Chinese setting with automatically generated distractors, while CC-Riddle (Xu et al., 2022) compiles Chinese character riddles via web crawling, language-model generation, and manual filtering. More recent work, such as RISCORE (Panagiotopoulos et al., 2025), uses automated prompting to reconstruct contextualized sentence puzzles and improve in-context riddle solving. Overall, these datasets remain confined to a few well-resourced languages, with non-English components typically smaller or partially synthetic.

A parallel line of work evaluates LLMs directly on riddles and game-like language tasks. Manna et al. (2024) study LLM performance on word-based games, including riddles, probing whether models leverage lexical and semantic structure beyond surface pattern matching. In educational contexts, Benelrhali and Berrada (2025) examines how AI tools and LLMs scaffold students' logical reasoning in riddle-based activities. At the group level, Rademaker et al. (2025) analyzes collaborative "Black Story" riddle solving, characterizing cognitive and interactional dynamics that current models do not yet capture.

Multilingual and culturally grounded riddle reasoning has only recently begun to receive attention, and work on genuinely low-resource languages remains sparse. Saxena et al. (2025) introduces the *Riddle of Reflection*, using Indian riddles to probe reasoning and self-awareness in multilingual LLMs and showing sharp variation across languages. CC-Riddle (Xu et al., 2022) exemplifies a broader pattern in non-English settings, where riddle data are collected via web crawling and language-model generation and then lightly filtered, yielding semi-synthetic resources with narrow coverage. On the generative side, Le et al. (2025) propose adaptive originality filtering and rejection-based prompting with a RiddleScore metric for multilingual riddle generation, but their data are synthetic, limited to a few language pairs, and relatively small per language. Riddle-style reasoning has also been extended to multimodal domains. Ghosal et al. (2025) introduce AlgoPuzzleVQA, a benchmark of visual–textual algorithmic puzzles that stress-test multimodal reasoning, again in settings where textual content is drawn from high-resource languages.

As a result, there is still no widely adopted, systematically constructed, human-authored riddle dataset for Bangla, despite its status as a major world language with a rich oral riddle tradition and relatively limited NLP resources. BANGLARIDDLEEVAL addresses this gap as a multi-task benchmark targeting Bangla riddles, built from diverse, culturally grounded human-authored riddles rather than synthetic prompts. It is designed to probe answer accuracy, reasoning quality, and robustness to carefully constructed distractors in a low-resource language setting.

## 3 Corpus Creation

The creation of BANGLARIDDLEEVAL involved a multi-stage process designed to transform raw folk riddles into a structured benchmark for assessing the reasoning capabilities of LLMs in Bengali.

### 3.1 Dataset Collection

The core dataset consists of Bengali riddles collected from human curated riddle books: ১০০০ ধাঁধা[1], ধাঁধা মানেই বাঁধা[2], and মজার ধাঁধা[3]. We first captured high-resolution photos of the relevant pages from these books, and then applied optical character recognition (OCR) using the `Gemini-2.5-Flash` model to convert the scanned text into a structured JSON format. The raw collection initially contained duplicates, OCR noise, and varying formats. We performed rigorous data cleaning to remove exact duplicates, correct obvious OCR errors, and normalize the text script. This process resulted in a consolidated set of unique riddles (stored in `riddles.json`), where each entry comprises a unique identifier (`riddle_id`), the riddle text (`riddle`), and the ground-truth answer (`ans`).

### 3.2 Synthetic Artifacts Generation

To transform the raw riddles into a reasoning benchmark, we enriched the data with three distinct layers of annotation: step-by-step reasoning, multiple-choice distractors, and semantic ambiguity resolution.

However, obtaining high-quality synthetic supervision at scale is challenging. Human-authored artifacts are costly and slow to produce for large datasets (Gierl et al., 2017; Zellers et al., 2019), and purely manual construction can introduce spurious correlations or annotation artifacts (Gururangan et al., 2018; Poliak et al., 2018; Tu et al., 2020). Our goal, therefore, is to automatically generate artifacts (reasoning chains, distractors, ambiguity annotations) that (i) remain semantically faithful to the riddle and gold answer, (ii) are close enough to the correct answer to be plausible and adversarial without becoming another valid answer (Goodrich, 1977), and (iii) are robust to candidate-only bias, discouraging models from solving items purely from the options or auxiliary signals (Si et al., 2019; Yu et al., 2020).

For the **Reasoning Annotation**, which is stored in `riddles_reasoning.json`, we utilized a high-capability LLM (GPT-4o) to generate Chain-of-Thought (CoT) explanations. To ensure quality, we employed an iterative prompting strategy where the model was provided with both the question and the answer. The generation process underwent three iterations to refine the logic, adhering to a strict four-step schema: (1) Answer Identification, (2) Metaphor Explanation, (3) Connection to Answer, and (4) Conclusion. This resulted in high-fidelity explanations that bridge the gap between the abstract riddle and the concrete answer.

For the **MCQ Generation**, stored in `riddles_mcq.json`, we construct four options for each riddle: one correct answer and three plausible but incorrect distractors. The MCQs are generated via GPT-4o, with special attention to ensuring that distractors are semantically related to the riddle's domain (e.g., other fruits if the answer is a fruit), which we enforce through prompt constraints and manual review of a sampled subset.

---

[1] https://www.rokomari.com/book/63201/1000-dhada
[2] https://www.rokomari.com/book/26795/dhadha-manei-badha
[3] https://www.rokomari.com/book/14553/mojar-dhadha

| Task | Question | Gemini Generated Answer | Correct Answer |
|---|---|---|---|
| Generative Question Answering | বন থেকে বেরুল টিয়ে সোনার টোপর মাথায় দিয়ে। *(The parrot came out of the forest wearing a golden crown on its head.)* Solve this Riddle. | আনারস *(Pineapple)* ✓ | আনারস *(Pineapple)* |
| Multiple Choice Question | এই ঘরে যাই, ওই ঘরে যাই দুম দুমিয়ে খাই। *(When I go to this room and that room, I'm slammed and banged around.)* Choose the correct option: A. ঘাঁটা *(Stirring stick)* B. কড়াই *(Frying pan)* C. বাটি *(Bowl)* D. চুলা *(Stove)* | B. কড়াই *(Frying pan)* ✗ | A. ঘাঁটা *(Stirring stick)* |
| Semantic Ambiguity Resolution | হাত আছে, পা নেই, বুক তার ফাটা মানুষকে গিলে খায়, নাই তার মাথা। *(It has hands but no feet, its chest is split open, it swallows a person, and it has no head.)* In this riddle, what does the word 'hand' refer to? A. আসল মানুষের হাত *(A real human hand)* B. হাতের আঙুল *(Fingers of the hand)* C. শার্টের হাতা *(Sleeve of a shirt)* D. হাতের গ্রিপ *(Hand grip)* | C. শার্টের হাতা *(Sleeve of a shirt)* ✓ | C. শার্টের হাতা *(Sleeve of a shirt)* |
| Explanation/Reasoning | Riddle: অন্ধকার ঘরে বাঁদরবা নাচে না না করলে আরও নাচে। *(In a dark room, a monkey dances; if you say 'no, no', it dances even more.)* Answer: জিহ্বা *(Tongue)*. Given a riddle and its correct answer, generate a concise four-step explanation, explicitly following (1) Answer Identification, (2) Metaphor Explanation, (3) Connection to Answer, and (4) Conclusion, that logically links the riddle's imagery to the answer. | ...The "dark room" is the mouth cavity. The "monkey dancing" refers to the constant, agile movement of the tongue while speaking... ✓ | ...The phrase "in a dark room" metaphorically refers to the inside of the mouth—a dark, closed space. The "monkey" here metaphorically represents the tongue... |

Table 1: Task overview of BANGLARIDDLEEVAL. Question, LLM-generated answer, and ground truth answer for four Bangla riddle tasks. Green ticks (✓) and red crosses (✗) indicate correctness of the generated answers.

Finally, for **Semantic Ambiguity Resolution**, stored in `riddles_semantic_ambiguity.json`, we annotated the specific "distractor word" or metaphorical trigger within each riddle. This annotation isolates the specific linguistic capability required to solve the riddle. For this task, we leverage GPT-4o to first automatically detect the ambiguous trigger word in the riddle and then generate four closely related candidate options that encode possible intended meanings of that word (e.g., alternative senses, referents, or glosses). Among these options, exactly one corresponds to the correct intended meaning, while the remaining three are incorrect but semantically close. The model is required to select the single option that correctly captures the intended meaning in context, thereby resolving the underlying semantic ambiguity.

### 3.3 Overview of Tasks

We formulate four distinct evaluation tasks within BANGLARIDDLEEVAL to test different facets of LLM capability, ranging from generative creativity to logic identification, as shown in Table 1.

**Task 1: Generative Question Answering (Generative QA)** This is the fundamental task of evaluating the model's ability to solve the riddle in an open-ended setting. The model is provided with the riddle text (Input) and must generate the correct entity (Output). This task assesses the model's retrieval of cultural knowledge and direct problem-solving ability.

**Task 2: Multiple Choice Question (MCQ)** In this task, the model is presented with the riddle and four options (one correct answer and three distractors). The model must select the correct option. This discriminative task is generally easier than the generative task but specifically tests the model's robustness against plausible distractors. To prevent models from exploiting positional cues, the options are randomized for every instance during evaluation.

**Task 3: Semantic Ambiguity Resolution (Logic Task)** This **novel task** is designed to test deep semantic and metaphorical understanding. The Bengali riddles rely on a "distractor word" (ambiguous trigger), used metaphorically to mislead the reader. In this task, we highlight the trigger word and present four closely related candidate meanings (e.g., alternative senses, referents, or glosses). Exactly one option corresponds to the correct intended meaning within the riddle, while the remaining three are incorrect but semantically close. The model must select the single option that best captures the intended meaning in context, thereby resolving the underlying semantic ambiguity and demonstrating sensitivity to figurative language rather than surface-level cues.

**Task 4: Riddle Explanation (Reasoning)** This task evaluates the model's CoT capabilities. The model is provided with the riddle and the correct answer, and it must generate a step-by-step explanation justifying why the answer fits the riddle. We evaluate the generated explanation against our reference reasoning dataset (from `riddles_reasoning.json`). This tests whether the model hallucinated the logic, even if it guessed the answer correctly.

### 3.4 Dataset Statistics

The final BANGLARIDDLEEVAL corpus is a unified collection of exactly 1,244 unique riddles, each instantiated across all four evaluation tasks, resulting in a total of **4,976** artifacts. This design yields a balanced benchmark in which every riddle is si-multaneously tested for generative capability, reasoning depth, distractor robustness, and ambiguity resolution. The dataset exhibits rich linguistic diversity, with an average riddle length of approximately 69 characters (about 13 words). A detailed breakdown of per-task statistics, along with the overall count of 4,976 artifacts, is presented in Table 2.

| Task | Samples | Avg Input Words | Avg Output Words |
| --- | --- | --- | --- |
| Generative QA | 1,244 | 13.48 | 1.45 |
| Reasoning | 1,244 | 13.48 | 206.99 |
| MCQ | 1,244 | 13.48 | 1.43 |
| Ambiguity Resolution | 1,244 | 13.48 | 3.90 |
| **Total Artifacts** | **4,976** | – | – |

Table 2: Detailed statistics of the BANGLARIDDLEEVAL benchmark. All tasks cover the full set of 1,244 riddles.

The reasoning task remains the most computationally demanding: explanations average over 200 words, expanding the original riddle context by a factor of roughly 16×. This substantial expansion enables evaluation not only of answer correctness but also of a model's ability to construct coherent, multi-step reasoning chains in Bangla.

## 4 Experimental Setup

### 4.1 Baseline Models

We evaluate BANGLARIDDLEEVAL on a diverse suite of large language models covering both open-source and proprietary systems, and spanning small to medium parameter scales. Our open-source baselines include GPT-OSS-20B (Agarwal et al., 2025), DeepSeek-R1-7B, DeepSeek-R1-14B (Guo et al., 2025), Qwen3-4B, Qwen3-8B, Qwen3-14B (Yang et al., 2025), Gemma3-4B, and Gemma3-12B (Team et al., 2025). As a strong proprietary reference, we additionally evaluate Gemini-2.5-Flash. We use the publicly available instruction-tuned variants with default decoding settings recommended by their providers. This collection allows us to probe how model family, size, and training regime affect performance on Bangla riddle reasoning.

### 4.2 Evaluation Metrics

We use task-specific metrics tailored to each component of BANGLARIDDLEEVAL. For the Multiple Choice (MCQ) and Semantic Ambiguity Resolution tasks, we report simple accuracy over the discrete option set. For the Generative QA task, we use BERTScore (Zhang* et al., 2020) to capture the semantic similarity between model outputs and gold answers, and additionally rely on an LLM-as-a-Judge (Badshah and Sajjad, 2025) to assess semantically correct but lexically divergent answers. For the reasoning task, we again use the LLM-as-a-Judge to assess the coherence, logical soundness, and faithfulness of the reasoning traces to the riddle and its ground-truth answer.

For the LLM-as-a-Judge component, we use Gemini-2.5-Flash as an automatic evaluator. Given the riddle, the gold answer, and the model's prediction, the judge is prompted to decide whether the prediction is correct for the Generative QA task, and to output a binary label (yes/no). We then compute the accuracy as the percentage of yes judgments. For the Reasoning task, the judge is instead instructed to assign a scalar score in the range [0, 10] along with a brief justification.

To validate the reliability of the LLM-as-a-Judge, we conducted a focused human review with a subject-matter expert (SME) fluent in Bangla and familiar with riddle reasoning. We designed a 0–100 human scoring rubric covering three components: (i) **Equivalence decision accuracy** (0–50 points), which asks whether the judge made the correct YES/NO decision about semantic equivalence; (ii) **Reasoning alignment** (0–40 points), which evaluates whether the analysis is logical and correctly supports the decision; and (iii) **Explanation clarity and justification** (0–10 points), which measures how clear and well-justified the explanation is. The SME scored 50 output samples using this rubric and ensured uniformity of judgments. The LLM-as-a-Judge achieved a Human Consistency Score ranging from 92.71 to 95.46 out of 100 across experiments, indicating that its decisions align closely with expert reasoning. Minor discrepancies typically arose from explanations that were essentially correct but slightly underspecified. Overall, this analysis supports the reliability of the judge's scores as a proxy for expert human evaluation.

### 4.3 Prompting Techniques

We systematically probe model behavior under three prompting regimes: zero-shot, few-shot, and chain-of-thought (CoT). In the *zero-shot* setting, the model receives only a concise task description together with the riddle (and, for MCQ-style tasks, the answer options), and must produce an answer without any in-context examples. In the *few-shot* setting, we prepend three example riddles with gold answers (and explanations, where appro-priate) randomly sampled from the dataset, explicitly excluding the instance being answered. This setup tests the model's in-context learning capabilities. In the *CoT* setting, we instruct the model to "think step by step" and provide exemplars with intermediate reasoning, encouraging it to generate explicit reasoning traces before committing to a final answer. For the detailed prompt settings, refer to the Appendix A.

### 4.4 Human Performance Evaluation

To contextualize model performance on the multiple-choice setting, we also estimate human accuracy on the MCQ components. We recruit three native Bangla-speaking undergraduate students, who independently answer 100 riddles randomly sampled from the dataset. Participants are instructed not to search for answers online and not to use any language models during the evaluation period, ensuring that their responses reflect their own riddle comprehension. For each annotator, we compute accuracy on the sampled riddles and then average these scores to obtain a single human performance estimate. This human baseline serves as a reference point for interpreting model results, indicating how close current LLMs are to competent native speakers on Bangla riddle tasks.

## 5 Result & Analysis

In this section, we present our results and findings on BANGLARIDDLEEVAL, organizing our discussion around the five research questions (RQs).

### 5.1 RQ1: To what extent can current LLMs solve Bangla riddles across the direct generative question answering task?

|  | BS_F1 | | | LLM Acc. (%) | | |
|---|---|---|---|---|---|---|
| Model | Z | F | C | Z | F | C |
| GPT-OSS-20B | 0.750 | 0.745 | 0.701 | 20.40 | 23.60 | **23.47** |
| DeepSeek-R1-7B | 0.745 | 0.747 | 0.693 | 2.33 | 2.13 | 4.40 |
| DeepSeek-R1-14B | 0.763 | 0.772 | 0.652 | 6.07 | 5.93 | 13.27 |
| Qwen3-4B | 0.738 | 0.715 | 0.690 | 9.07 | 5.40 | 10.67 |
| Qwen3-8B | 0.772 | 0.768 | <u>0.708</u> | 11.13 | 10.53 | 11.53 |
| Qwen3-14B | 0.781 | 0.782 | **0.730** | 16.80 | 15.20 | <u>14.87</u> |
| Gemma3-4B | 0.779 | 0.780 | 0.645 | 10.20 | 8.27 | 9.40 |
| Gemma3-12B | <u>0.786</u> | <u>0.787</u> | 0.668 | 19.47 | 17.87 | 13.13 |
| Gemini-2.5-Flash | **0.807** | **0.814** | 0.698 | **29.13** | **28.67** | 11.87 |

Table 3: Generative QA results on Bangla riddles. BS_F1 = BERTScore$_{F1}$, LLM Acc. = LLM-as-a-Judge accuracy (%). Z = Zero-Shot, F = Few-Shot, C = CoT. Best scores are in bold, second-best are underlined.

The results shown in Table 3 indicate that across nine LLMs and three prompting regimes, we observe moderate lexical–semantic overlap but relatively low semantic correctness on Bangla riddles. BERTScore$_{F1}$ remains in a narrow band around 0.74–0.81 for zero-shot and few-shot prompts, with CoT often slightly worse (e.g., best zero-shot 0.807 vs. best CoT 0.730), indicating that models can usually produce surface-similar answers but struggle with precise riddle resolution. The LLM-as-a-Judge accuracy is substantially lower, ranging from roughly 2–29%, and again shows no consistent benefit from CoT prompting–indeed, the strongest generative regime is often zero-shot or few-shot (e.g., up to 29.13% vs. CoT 23.47%)–suggesting that, while current LLMs capture some semantic cues, fully "solving" Bangla riddles in a generative setting remains a very challenging task for LLMs. In other words, even the best model configuration correctly answers only about one in four riddles, implying that current LLMs can handle Bangla riddles to a limited extent but are still far from robust, human-like performance.

### 5.2 RQ2: Are LLMs robust to semantically plausible distractors in Bangla riddle MCQs, and how close are they to human performance?

| Model | Zero-Shot | Few-Shot | CoT |
|---|---|---|---|
| GPT-OSS-20B | 41.49 | 40.43 | 44.49 |
| DeepSeek-R1-7B | 24.76 | 28.36 | 39.34 |
| DeepSeek-R1-14B | 36.52 | 38.89 | 42.20 |
| Qwen3-4B | 40.45 | 38.12 | 28.57 |
| Qwen3-8B | 34.01 | 32.09 | 33.33 |
| Qwen3-14B | <u>42.70</u> | <u>42.01</u> | <u>44.55</u> |
| Gemma3-4B | 27.33 | 32.42 | 18.75 |
| Gemma3-12B | 27.00 | 27.95 | 31.80 |
| Gemini-2.5-Flash | **52.33** | **54.00** | **56.33** |
| Human (MCQ) | | 83.00 | |

Table 4: MCQ accuracy (%) under different prompting strategies. Random guessing over four options corresponds to 25% accuracy. Bold indicates the best model per column; underlined indicates the second best.

The results shown in Table 4 indicate that across all models, the accuracy remains far below the 83% human baseline, even for the strongest system (Gemini-2.5-Flash), which peaks at 56.33% under CoT prompting. Mid-sized open-source models (e.g., Qwen3-14B, GPT-OSS-20B) typically cluster in the low-to-mid 40% range. In contrast, smaller or less capable models often hover near or only slightly above the 25% random baseline, especially in zero-shot mode. This gap indicates that semantically plausible distractors are effective: models cannot reliably exploit simple candidate-only heuristics and often fail to distinguish the correct answer from closely related alternatives. CoT prompting tends to yield modest improvements for several models (e.g., DeepSeek-R1-7B, DeepSeek-R1-14B, Gemini-2.5-Flash), but these gains are insufficient to close the substantial gap to human performance, suggesting that robust Bangla riddle understanding under adversarial distractors remains an open challenge for current LLMs.

### 5.3 RQ3: How well do LLMs generate faithful and coherent step-by-step explanations for Bangla riddles?

We evaluate explanation quality using the LLM-as-a-Judge score in the range [0, 10], where higher scores indicate more faithful, coherent, and well-justified reasoning.

| Model | LLM-as-a-Judge (0–10) |
| --- | --- |
| GPT-OSS-20B | <u>8.28</u> |
| DeepSeek-R1-7B | 0.87 |
| DeepSeek-R1-14B | 2.82 |
| Qwen3-4B | 4.38 |
| Qwen3-8B | 5.25 |
| Qwen3-14B | 6.47 |
| Gemma3-4B | 6.17 |
| Gemma3-12B | 7.98 |
| Gemini-2.5-Flash | **8.71** |

Table 5: Reasoning quality on BANGLARIDDLEEVAL, measured by LLM-as-a-Judge (0–10). Bold indicates the best model; underlined indicates the second best.

As shown in Table 5, reasoning quality varies substantially across models. The strongest proprietary model, Gemini-2.5-Flash, attains the highest score (8.71), followed by GPT-OSS-20B (8.28) and Gemma3-12B (7.98), indicating that these systems can often produce coherent and faithful step-by-step explanations in Bangla. Mid-sized open-source models such as Qwen3-14B and Gemma3-4B achieve moderate scores in the 6–6.5 range, suggesting partially correct but less consistently well-structured reasoning. In contrast, some models with comparable or even larger capacity (e.g., DeepSeek-R1-7B, DeepSeek-R1-14B) score poorly (below 3), frequently failing to provide meaningful or logically aligned explanations, likely reflecting differences in training data or multilingual alignment rather than raw parameter count alone. Overall, high-end models demonstrate reasonably strong explanatory abilities, but there remains a clear gap across the model spectrum in generating high-quality Bangla riddle reasoning.

### 5.4 RQ4: How well do LLMs select the intended meaning of ambiguous trigger words in Bangla riddles?

| Model | Zero-Shot | Few-Shot | CoT |
| --- | --- | --- | --- |
| GPT-OSS-20B | <u>65.53</u> | 63.54 | **65.77** |
| DeepSeek-R1-7B | 30.77 | 31.33 | 26.03 |
| DeepSeek-R1-14B | 55.78 | 58.98 | 55.78 |
| Qwen3-4B | 56.94 | **65.19** | 44.44 |
| Qwen3-8B | 60.20 | 61.15 | 57.14 |
| Qwen3-14B | 64.00 | 63.21 | <u>62.92</u> |
| Gemma3-4B | 37.67 | 40.62 | 31.25 |
| Gemma3-12B | 55.67 | 54.44 | 26.00 |
| Gemini-2.5-Flash | **67.67** | 63.33 | 60.67 |

Table 6: Semantic Ambiguity Resolution accuracy (%) on BANGLARIDDLEEVAL under different prompting strategies. Random guessing over four options corresponds to 25% accuracy. Bold indicates the best model per column; underlined indicates the second best.

The results shown in Table 6 indicate that across models, accuracy on the Semantic Ambiguity Resolution task ranges from roughly 26% to 68%, indicating that most LLMs perform clearly above the 25% random baseline but still far from perfectly resolving word-level ambiguity. Stronger models such as Gemini-2.5-Flash and GPT-OSS-20B achieve the best zero-shot accuracies (67.67% and 65.53%, respectively), suggesting that large models can often recover the intended sense of ambiguous trigger words. Mid-sized open-source models (e.g., Qwen3-8B, Qwen3-14B, DeepSeek-R1-14B) consistently reach the mid–50%s to low–60%s, while smaller or weaker models (e.g., DeepSeek-R1-7B, Gemma3-4B) hover near or slightly above chance. Prompting effects are mixed: few-shot prompting notably boosts Qwen3-4B (from 56.94% to 65.19%), but CoT sometimes hurts performance (e.g., Qwen3-4B, Gemma3-12B), indicating that explicit reasoning prompts do not uniformly benefit lexical disambiguation. Overall, LLMs show moderate competence at selecting the intended meaning of ambiguous trigger words in Bangla riddles, but substantial headroom remains.

### 5.5 RQ5: How do model choice (family, size, openness) and prompting strategy (zero-shot, few-shot, CoT) affect LLM performance across Bangla riddle tasks?

To obtain a single comparable score per model and prompting regime, we first normalize each metric to the [0, 1] range across all systems: MCQ accuracy, Generative QA BERTScore, Generative QA LLM-as-a-Judge, and Semantic Ambiguity Reso-lution accuracy (plus Reasoning LLM-as-a-Judge for CoT only). For each model and prompting strategy, we then average the available normalized metrics, yielding three aggregate scores per model (Zero-Shot, Few-Shot, CoT).

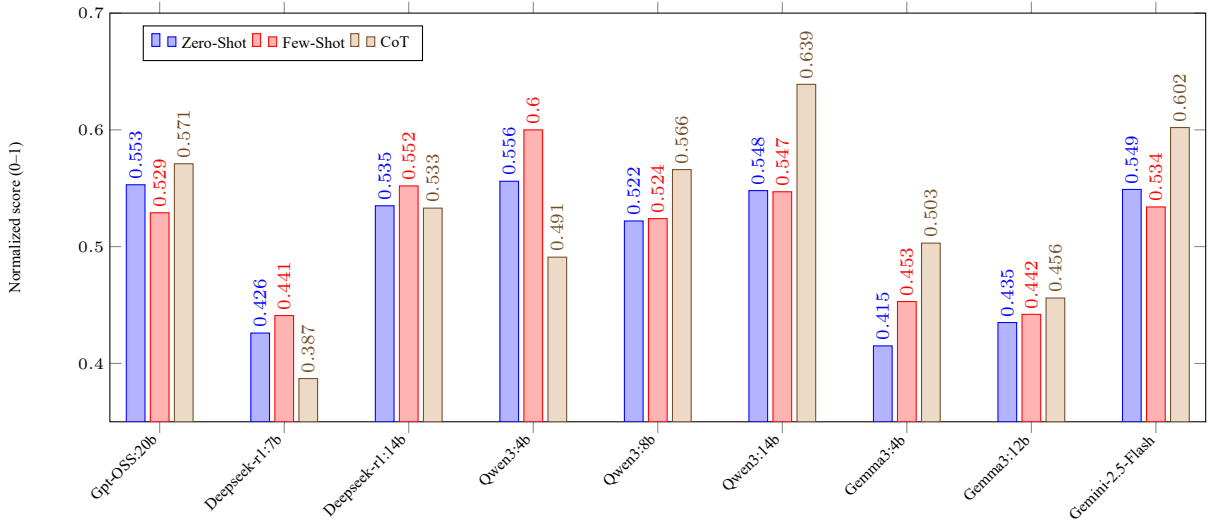

Figure 2: Aggregate normalized performance of each model across prompting regimes on BANGLARIDDLEEVAL.

Figure 2 summarizes the aggregate normalized performance of each model across all tasks and prompting regimes. Overall scores cluster in a moderate band (roughly 0.40–0.65), with clear variation by model family and size. The strongest proprietary model (Gemini-2.5-Flash) and the largest open-source models (Qwen3:14B and Gpt-OSS:20B) consistently occupy the top tier, while smaller or weaker models (e.g., Deepseek-r1:7B, Gemma3:4B) lag behind across all prompts. Prompting effects are present but secondary: CoT tends to yield the best aggregate scores for most higher-capacity models (Gpt-OSS:20B, Qwen3:14B, Gemma3:12B, Gemini-2.5-Flash), whereas few-shot prompting helps some weaker models (e.g., Deepseek-r1:7B, Qwen3:4B). Zero-shot remains a strong baseline for several systems but is rarely optimal. Taken together, these trends suggest that model choice is the primary driver of performance on Bangla riddles, with prompting strategy providing incremental gains, especially for already strong models.

## 6 Discussion

Across tasks and models, several consistent patterns emerge from BANGLARIDDLEEVAL. First, Bangla riddles are genuinely hard for current LLMs: even the best systems achieve only modest LLM-as-a-Judge accuracy on generative QA and remain far below the 83% human baseline on MCQs, despite producing reasonably high BERTScore values. Our hard, semantically plausible distractors and metaphor-heavy prompts successfully stress-test reasoning rather than surface similarity. Second, while larger and better-trained families consistently outperform smaller or less aligned models, prompting strategies play a secondary role: zero-shot is often competitive, few-shot helps selectively, and CoT yields gains for some strong models but can hurt lexical disambiguation. Third, explanation and ambiguity-focused evaluations reveal deeper weaknesses: only top-tier models reach high judge scores on step-by-step reasoning and reliably pick the intended sense of ambiguous trigger words. Overall, the results suggest that current LLMs capture some of the semantic and cultural cues needed for Bangla riddle reasoning, but still lack robust, human-like competence, underscoring the need for stronger multilingual training and Bangla reasoning benchmarks.

## 7 Conclusion

We introduce BANGLARIDDLEEVAL, the first benchmark to systematically test LLMs on Bangla riddle reasoning across four complementary tasks. Our results show that, while current models capture some semantic and cultural cues, they still lag far behind humans, especially under hard distractors and semantic ambiguity. BANGLARIDDLEEVAL thus offers a challenging benchmark for pushing multilingual LLMs toward more robust, figurative, and low-resource reasoning.

# Limitations

While BANGLARIDDLEEVAL is a first systematic benchmark for Bangla riddle reasoning, it is not exhaustive. The riddles are drawn from a limited set of printed books, which may under-represent regional, dialectal, or contemporary variants of Bangla riddles. As a result, our dataset may skew toward certain styles and difficulty levels rather than fully capturing the breadth of Bangla riddling traditions. Moreover, a substantial part of the supervision signal (explanations, distractors, and ambiguity options) and our LLM-as-a-Judge evaluation rely on large proprietary models (e.g., GPT-4o, Gemini-2.5-Flash). Despite quality checks and SME validation, these components can introduce modeling biases and occasional annotation errors. Our empirical analysis is also restricted to a specific set of models and prompting strategies, so the reported results should be viewed as indicative rather than definitive upper bounds on Bangla riddle reasoning.

# Ethical Considerations

All riddle data in BANGLARIDDLEEVAL are sourced from publicly available Bangla riddle books and online resources, used strictly for non-commercial, research purposes, with appropriate attribution. No personal or sensitive information about individuals is contained in the dataset. For the human studies, we recruited native Bangla-speaking undergraduate volunteers and one subject-matter expert (SME) for evaluation and rubric validation. Participation was fully voluntary, with informed consent obtained prior to data collection, and no identifying information was stored or reported. All experimental procedures were designed to pose minimal risk, aligning with standard ethical practices for NLP and human evaluation studies.

## A Appendix

### A.1 Prompting Hyperparameters

To reduce confounding factors, we keep decoding hyperparameters fixed across models and prompting regimes, varying only the prompt format (zero-shot, few-shot, CoT). All models use greedy-ish decoding with low temperature, moderate nucleus sampling, and a shared maximum generation length; CoT settings use a slightly larger budget to allow for longer reasoning traces.

| Hyperparam | Zero-Shot | Few-Shot | CoT |
|---|---|---|---|
| Temp | 0.3 | 0.3 | 0.3 |
| Top-$k$ | 40 | 40 | 40 |
| Top-$p$ | 0.95 | 0.95 | 0.95 |
| MaxTok-QA/MCQ/SA | 64 | 64 | 64 |
| MaxTok-Reasoning | – | – | 1024 |
| #Shots | – | 3 | – |
| Out-lang | bn | bn | bn |
| CoT-lang | – | – | bn (RE), en (QA, MCQ, SA) |

Table 7: Summary of decoding hyperparameters across zero-shot, few-shot, and CoT settings. Here: Temp = temperature, QA/MCQ/SA/RE = Generative QA/Multiple Choice Questions/Semantic Ambiguity/Reasoning tasks, MaxTok-QA/MCQ/SA = max new tokens for QA/MCQ/SA, MaxTok-Reasoning = max new tokens for reasoning, Out-lang = output language, en = English, bn = Bangla.

### A.2 Prompt Templates

Below, we provide the core prompt templates used in our experiments for zero-shot, few-shot, and chain-of-thought (CoT) prompting across all tasks (Generative QA, MCQ, Semantic Ambiguity), as well as the LLM-as-a-Judge and Bangla reasoning prompts. These templates are reproduced verbatim from our implementation.

---

**Zero-shot Prompt: All Tasks**

```
PROMPT_ALL_TASKS_ZERO_SHOT = (
"You are an AI assistant that solves
Bengali riddles. [Generative QA]\n"
"You are an AI assistant that answers
Bengali riddle multiple-choice
questions. [MCQ]\n"
"You are an AI assistant that analyzes
Bengali riddle semantic ambiguity
questions. [Semantic Ambiguity]\n"
"Task:\n"
"1. Read the Bengali riddle: {riddle}
[All Tasks]\n"
"2. Read the semantic question:
{question} [Semantic Ambiguity]\n"
"3. Review the provided answer choices:
{options} [MCQ, Semantic Ambiguity]\n"
"4. Provide the most accurate answer in
Bengali. [Generative QA]\n"
"5. Select the **single most accurate
answer** and respond in Bengali.
[MCQ]\n"
"6. Select the **single most
accurate answer** that best explains
the semantic meaning. [Semantic
Ambiguity]\n\n"
"Response Rules:\n"
"- Provide ONLY the answer in Bengali
text [Generative QA]\n"
"- Answer must be a single Bengali word
or short Bengali phrase [Generative
QA]\n"
"- Do NOT use English words or
explanations [Generative QA]\n"
"- The index must be the programming
list index (starting from 0). [MCQ,
Semantic Ambiguity]\n"
"- Respond ONLY with the exact format
below. [MCQ, Semantic Ambiguity]\n"
"- Use Bengali text for the answer
option. [MCQ, Semantic Ambiguity]\n"
"- Do NOT add explanations, reasoning,
or extra text [All Tasks]\n"
"- Respond with ONLY the Bengali answer
[All Tasks]\n"
"- Follow this exact structure: [MCQ,
Semantic Ambiguity]\n"
"Index: <option_index>, Answer:
<option_text_in_Bengali>[MCQ, Semantic
Ambiguity]"
)
```

---

**Few-shot Prompt: All Tasks**

```
PROMPT_FEW_SHOT_ALL = (
"You are an AI assistant that analyzes
Bengali riddle semantic ambiguity
questions. [Semantic Ambiguity]\n"
"You are an AI assistant that answers
Bengali riddle multiple-choice
questions. [MCQ]\n"
"You are an AI assistant that solves
Bengali riddles. [Generative QA]\n"
"Task:\n"
"1. Read the original riddle: {riddle}
[Semantic Ambiguity]\n"
"1. Read the riddle in Bengali:
{riddle} [MCQ]\n"
"Task: Read the Bengali riddle below
and provide the most accurate answer in
Bengali. [Generative QA]\n"
"2. Read the semantic question:
{question} [Semantic Ambiguity]\n"
"2. Review the provided answer choices:
{options} [MCQ, Semantic Ambiguity]\n"
"3. Select the **single most accurate
answer** and respond in Bengali.
[MCQ]\n"
"4. Select the **single most
accurate answer** that best explains
the semantic meaning. [Semantic
Ambiguity]\n\n"
"Response Rules: [MCQ, Semantic
Ambiguity]\n"
"- The index must be the programming
list index (starting from 0). [MCQ,
```

```
Semantic Ambiguity]\n"
"- Respond ONLY with the exact format
below. [MCQ, Semantic Ambiguity]\n"
"- Use Bengali text for the answer
option. [MCQ, Semantic Ambiguity]\n"
"- Do NOT add explanations, extra words,
reasoning steps, or anything outside
the specified format. [MCQ, Semantic
Ambiguity]\n"
"- Follow this exact structure: [MCQ,
Semantic Ambiguity]\n"
"Index: <option_index>, Answer:
<option_text_in_Bengali>[MCQ, Semantic
Ambiguity]\n\n"
"Now answer for the given semantic
question. [Semantic Ambiguity]\n"
"Now answer for the given riddle.
[MCQ]\n"
"Now solve this riddle following the
same format: [Generative QA]\n"
"Riddle: {riddle} [Generative QA]\n\n"
"Response Requirements: [Generative
QA]\n"
"- Provide ONLY the Bengali answer like
the examples above [Generative QA]\n"
"- Do NOT add any explanations,
reasoning, or extra text [Generative
QA]\n"
"- Answer must be a single Bengali word
or short Bengali phrase [Generative
QA]\n"
"- Do NOT use English words or
explanations [Generative QA]\n"
"- Follow the exact format: Answer:
<bengali_text> [Generative QA]\n\n"
"Answer: [Generative QA]"
)
```

### Chain-of-Thought Prompt: All Tasks

```
PROMPT_CHAIN_OF_THOUGHTS_ALL = (
"You are an AI assistant that solves
Bengali riddles using step-by-step
reasoning. [Generative QA]\n"
"You are an AI assistant that answers
Bengali riddle multiple-choice
questions using step-by-step reasoning.
[MCQ]\n"
"You are an AI assistant that analyzes
Bengali riddle semantic ambiguity
questions using step-by-step reasoning.
[Semantic Ambiguity]\n\n"
"Task:\n"
"1. Read the Bengali riddle below:
{riddle} [Generative QA]\n"
"1. Read the riddle in Bengali:
{riddle} [MCQ]\n"
"1. Read the original riddle: {riddle}
[Semantic Ambiguity]\n"
"2. Read the semantic question:
{question} [Semantic Ambiguity]\n"
"2. Review the provided answer choices:
{options} [MCQ]\n"
"3. Review the provided answer choices:
{options} [Semantic Ambiguity]\n"
"3. Select the **single most accurate
answer** using chain-of-thought
```

```
reasoning. [MCQ]\n"
"4. Select the **single most accurate
answer** using chain-of-thought
reasoning. [Semantic Ambiguity]\n"
"Provide the most accurate answer using
chain-of-thought reasoning. [Generative
QA]\n\n"
"Response Rules:\n"
"- Think step by step about the
riddle's metaphors and meaning
[Generative QA, MCQ]\n"
"- Think step by step about the
semantic ambiguity and metaphorical
meanings [Semantic Ambiguity]\n"
"- Write reasoning steps in English but
think with Bengali cultural knowledge
[All Tasks]\n"
"- Final answer MUST be in Bengali and
ONLY Bengali text [Generative QA]\n"
"- Final answer must be in Bengali [MCQ,
Semantic Ambiguity]\n"
"- The index must be the programming
list index (starting from 0). [MCQ,
Semantic Ambiguity]\n"
"- Be clear, concise, and factual
(avoid overly lengthy explanations).
[MCQ, Semantic Ambiguity]\n\n"
"In Reasoning_En, write step-by-step
reasoning in English --- break down the
solution logically:\n"
"Reasoning_En:\n"
"Step 1: <analyze the riddle's key
words and metaphors> [Generative QA]\n"
"Step 2: <connect observations with
possible meanings> [Generative QA]\n"
"Step 3: <eliminate wrong possibilities
with brief reasoning> [Generative
QA]\n"
"Step 4: <explain why the final choice
is correct> [Generative QA]\n\n"
"Final Answer:
<single_bengali_word_or_phrase>
[Generative QA]\n"
"Final Answer: Index: <option_index>,
Answer: <option_text_in_Bengali>[MCQ,
Semantic Ambiguity]"
)
```

### LLM-as-a-Judge: Generative QA

```
LLM_JUDGE_PROMPT_GENERATIVE = (
"You are an expert evaluator for
Bengali riddle answers. Your task
is to score how correct a predicted
answer is for a given riddle.\n\n"
"Riddle: {riddle}\n"
"Ground Truth Answer: {ground_truth}\n"
"Predicted Answer: {predicted}\n\n"
"Evaluation Criteria:\n"
"1. Exact Match: Are the answers
exactly the same?\n"
"2. Semantic Equivalence: Do they refer
to the same concept/object?\n"
"3. Cultural Context: Consider Bengali
cultural and linguistic variations\n"
"4. Acceptable Synonyms: Different
Bengali words for the same thing\n"
```

```
"5. Spelling Variations: Minor spelling
differences in Bengali\n"
"6. Partial Correctness: Consider
partial matches as valid (e.g., if
ground truth is '    ' and predicted is
'    ', give partial credit)\n"
"Scoring Instructions:\n"
"- Give a score between 0 and 1\n"
"- 1.0: Perfect match (exact or
semantically equivalent)\n"
"- 0.7-0.9: Very close match (minor
variations, synonyms, or slight
differences)\n"
"- 0.4-0.6: Partial match (partially
correct but missing important parts)\n"
"- 0.1-0.3: Poor match (completely
different but some tiny relevance)\n"
"- 0.0: Completely wrong or no
answer\n\n"
"Respond with ONLY the numerical score
(e.g., 0.8, 0.5, 1.0, 0.0)\n\n"
"Score:"
)
```

### LLM-as-a-Judge: Reasoning

```
LLM_JUDGE_PROMPT_REASONING = (
"You are an expert evaluator for
Bengali riddle reasoning explanations.
Your task is to score the quality of
a reasoning explanation for a given
riddle.\n\n"
"Riddle: {riddle}\n"
"Correct Answer: {ground_truth}\n"
"Generated Reasoning: {predicted}\n\n"
"Evaluation Criteria:\n"
"1. Logical Structure: Does the
reasoning follow a clear logical
progression?\n"
"2. Accuracy: Does the reasoning
correctly identify the metaphors and
connections?\n"
"3. Completeness: Does it cover the key
elements of the riddle?\n"
"4. Cultural Context: Does it
demonstrate understanding of Bengali
cultural context?\n"
"5. Language Quality: Is the Bengali
language clear and well-structured?\n"
"6. Conclusion: Does it reach
the correct answer through valid
reasoning?\n\n"
"Scoring Instructions:\n"
"- Give a score between 0 and 10\n"
"- 9-10: Excellent reasoning (logical,
accurate, complete, culturally
aware)\n"
"- 7-8: Good reasoning (mostly correct
with minor issues)\n"
"- 4-6: Average reasoning (some
correct elements but lacks depth or
accuracy)\n"
"- 1-3: Poor reasoning (limited
understanding, mostly incorrect)\n"
"- 0: No meaningful reasoning or
completely wrong\n\n"
"Respond with ONLY the numerical score
(e.g., 8, 5, 10, 0)\n\n"
"Score:"
)
```

### Bangla Reasoning Generation Prompt

```
BENGALI_REASONING_PROMPT = (
"আপনি একটি বাংলা ধাঁধার জন্য বিস্তারিত যুক্তি তৈরি
করবেন। (You will generate a detailed
reasoning for a Bangla riddle).\n\n"
"ধাঁধা:                    "{riddle}" (Riddle:
"{riddle}").\n"
"উত্তর:                     "{answer}" (Answer:
"{answer}").\n\n"
"নিচের ৪টি ধাপে বিশ্লেষণ করুন এবং একটি সুন্দর
বাংলা অনুচ্ছেদ আকারে লিখুন: (Analyze using
the four steps below and write a fluent
paragraph in Bangla).\n\n"
"1. উত্তর চিহ্নিতকরণ: ধাঁধার নির্দিষ্ট শব্দগুলো উদ্ধৃত করুন
(1. Answer identification: quote the
key words in the riddle).\n"
"2. রূপকের ব্যাখ্যা: রূপকটি কী প্রতিনিধিত্ব করে তা ব্যাখ্যা
করুন (2. Metaphor explanation: explain
what the metaphor stands for).\n"
"3.   উত্তরের সাথে সংযোগ:   {answer}-এর কোন
বৈশিষ্ট্য এই ধাঁধার সাথে মিলে যায় তা ব্যাখ্যা করুন
(3. Connection to the answer: explain
which properties of "{answer}" match
the riddle).\n"
"4. সিদ্ধান্ত: কেন এটাই একমাত্র যুক্তিসংগত উত্তর
তা সংক্ষেপে বলুন (4. Conclusion: briefly
justify why this is the only logically
consistent answer).\n\n"
"উদাহরণ ফরম্যাট: (Example format:).\n"
"১. 'এক থালা': এখানে আকাশকে একটি বিশাল থালার সাথে
তুলনা করা হয়েছে। (1. 'One plate': here the
sky is compared to a huge plate).\n"
"২. 'সুপারি': সুপারি যেমন ছোট ছোট গোল হয়, আকাশের
নক্ষত্রগুলোকেও দেখতে ছোট বিন্দুর মতো লাগে। (2.
'Betel nut': just as betel nuts are
small and round, the stars look like
tiny dots in the sky).\n"
"৩. 'গুনতে নারি': সুপারি গোনা সম্ভব হলেও, আকাশের
তারা বা নক্ষত্র অসংখ্য, যা গুনে শেষ করা যায় না।
(3. 'Cannot be counted': even if betel
nuts can be counted, the stars in the
sky are countless and cannot be fully
counted).\n"
"সিদ্ধান্ত: আকাশের বিশাল থালায় ছড়িয়ে থাকা অগণিত
নক্ষত্রই হলো এই ধাঁধার উত্তর। (Conclusion:
the countless stars scattered across
the vast sky are the answer to this
riddle).\n"
"অনুগ্রহ করে শুধুমাত্র reasoning টেক্সট দিন। JSON
বা অন্য কোনো ফরম্যাট ব্যবহার করবেন না। (Please
provide only the reasoning text. Do not
use JSON or any other format)."
)
```